\documentclass[10pt,twocolumn,letterpaper]{article}

\usepackage{wacv}
\usepackage{times}
\usepackage{epsfig}
\usepackage{graphicx}
\usepackage{amsmath}
\usepackage{amssymb}

\def\wacvPaperID{707} 

\wacvfinalcopy 

\ifwacvfinal
\def\assignedStartPage{9876} 
\fi

\ifwacvfinal
\usepackage[breaklinks=true,bookmarks=false]{hyperref}
\else
\usepackage[pagebackref=true,breaklinks=true,colorlinks,bookmarks=false]{hyperref}
\fi

\ifwacvfinal
\else
\fi

\begin{document}

\title{Attention-based Domain Adaptation for Single Stage Detectors}

\author{Vidit Vidit \;\;\; Mathieu Salzmann\\
CVLab,EPFL\\

{\tt\small firstname.lastname@epfl.ch}

}

\maketitle


\newif\ifdraft
\drafttrue

\newcommand{\bt}{\mathbf{t}}
\newcommand{\bM}{\mathbf{M}}
\newcommand{\br}{\mathbf{r}}
\newcommand{\bd}{\mathbf{d}}
\newcommand{\bX}{\mathbf{X}}
\newcommand{\bz}{\mathbf{z}}
\newcommand{\bG}{\mathbf{G}}
\newcommand{\bB}{\mathbf{B}}
\newcommand{\bS}{\mathbf{S}}
\newcommand{\bH}{\mathbf{H}}
\newcommand{\bR}{\mathbf{R}}
\newcommand{\bA}{\mathbf{A}}
\newcommand{\bdelta}{\boldsymbol{\delta}}

\newcommand{\addparagraphup}{\vspace*{0.01in}}

\newcommand{\ms}[1]{\ifdraft {\color{green}{#1}} \else {#1}\fi}
\newcommand{\vdt}[1]{\ifdraft {\color{blue}{#1}} \else {#1}\fi}

\newcommand{\MS}[1]{\ifdraft {\color{green}{\textbf{MS: #1}}}\else {}\fi}
\newcommand{\VDT}[1]{\ifdraft {\color{red}{\textbf{VDT: #1}}}\else {}\fi}

\newcommand{\comment}[1]{}
\newcommand{\teaserfig}[1]{./images/#1}


\iftrue 

\newcommand{\cutsectionup}{\vspace*{0in}}
\newcommand{\cutsectiondown}{\vspace*{-0.05in}}

\newcommand{\cutsubsectionup}{\vspace*{-0.05in}}
\newcommand{\cutsubsectiondown}{\vspace*{-0.04in}}

\newcommand{\cutsubsubsectionup}{\vspace*{-0.05in}}
\newcommand{\cutsubsubsectiondown}{\vspace*{-0.05in}}

\newcommand{\cutparagraphup}{\vspace*{-0in}}
\newcommand{\cutparagraphdown}{\vspace*{-0in}}

\newcommand{\cuthalfcaptionup}{\vspace*{-0.15in}}
\newcommand{\cuthalfcaptiondown}{\vspace*{-0.25in}}

\newcommand{\cutcaptionup}{\vspace*{-0.1in}}
\newcommand{\cutcaptiondown}{\vspace*{-0.15in}}

\newcommand{\cuthalftablecaptionup}{\vspace*{-0in}}
\newcommand{\cuthalftablecaptiondown}{\vspace*{-0.1in}}

\newcommand{\cuttablecaptionup}{\vspace*{-0in}}
\newcommand{\cuttablecaptiondown}{\vspace*{-0.1in}}

\newcommand{\cutequationup}{\vspace*{-0.05in}}
\newcommand{\cutequationdown}{\vspace*{-0.01in}}

\newcommand{\cuttableup}{\vspace*{-0.2in}}
\newcommand{\cuttabledown}{\vspace*{-0.3in}}

\newcommand{\cutabstractup}{\vspace*{-0.12in}}
\newcommand{\cutabstractdown}{\vspace*{-0.3in}}

\newcommand{\cutalgorithmup}{\vspace*{-0in}}
\newcommand{\cutalgorithmdown}{\vspace*{-0.1in}}

\newcommand{\cut}{{\vspace*{-0.02in}}}
\newcommand{\cutmore}{{\vspace*{-0.06in}}}
\newcommand{\negcut}{}
\else 
\newcommand{\cutsectionup}{}
\newcommand{\cutsectiondown}{}

\newcommand{\cutsubsectionup}{}
\newcommand{\cutsubsectiondown}{}

\newcommand{\cutsubsubsectionup}{}
\newcommand{\cutsubsubsectiondown}{}

\newcommand{\cutparagraphup}{}
\newcommand{\cutparagraphdown}{}

\newcommand{\cuthalfcaptionup}{}
\newcommand{\cuthalfcaptiondown}{}

\newcommand{\cutcaptionup}{}
\newcommand{\cutcaptiondown}{}

\newcommand{\cutequationup}{}
\newcommand{\cutequationdown}{}

\newcommand{\cuttableup}{}
\newcommand{\cuttabledown}{}

\newcommand{\cutabstractup}{}
\newcommand{\cutabstractdown}{}

\newcommand{\cutalgorithmup}{}
\newcommand{\cutalgorithmdown}{}

\newcommand{\cut}{}
\newcommand{\cutmore}{}
\newcommand{\negcut}{}
\fi


\begin{abstract}
   While domain adaptation has been used to improve the performance of object detectors when the training and test data follow different distributions, previous work has mostly focused on two-stage detectors. This is because their use of region proposals makes it possible to perform local adaptation, which has been shown to significantly improve the adaptation effectiveness. Here, by contrast, we target single-stage architectures, which are better suited to resource-constrained detection than two-stage ones but do not provide region proposals. To nonetheless benefit from the strength of local adaptation, we introduce an attention mechanism that lets us identify the important regions on which adaptation should focus. Our method gradually adapts the features from global, image-level to local, instance-level. Our approach is generic and can be integrated into any single-stage detector. We demonstrate this on standard benchmark datasets by applying it to both SSD and YOLOv5. Furthermore, for equivalent single-stage architectures, our method outperforms the state-of-the-art domain adaptation techniques even though they were designed for specific detectors.
\end{abstract}

\section{Introduction}
Modern object detection methods can be grouped into two broad categories: Two-stage architectures~\cite{ren2016faster}, that first extract regions of interest (ROIs) and then classify and refine them; and single-stage ones~\cite{liu2016ssd,glenn_jocher_2020_4154370,tian2019fcos}, that directly output bounding boxes and classes from the feature maps. While the former yield slightly higher accuracy, the latter are faster and more compact, making them better suited for real-time applications or for mobile devices.

In any event, all object detectors reach their best performance when the training and test data are acquired in the same conditions, such as using the same camera, in similar illumination conditions. When they are not, the resulting domain gap significantly degrades the detection results. Addressing this is the focus of domain adaptation~\cite{ganin2016domain,saito2018maximum,DBLP:conf/icml/LongC0J15,wang2019transferable,hoffman2018cycada,liu2018detach}. In this work, we focus on \emph{unsupervised} domain adaptation, whose goal is to bridge the gap between the source (training) and target (test) domain without having access to any target annotations.

The recent work on domain adaptation for object detection~\cite{chen2018domain,saito2019strong,chen2020harmonizing,zhu2019adapting,shen2019scl} has focused almost exclusively on two-stage detectors. At the heart of most of these methods lies the intuition that adaptation should be performed locally, focusing on the foreground objects because the background content may genuinely differ between the training and test data, whereas the object categories of interest do not. This process of local adaptation is facilitated by the ROIs used in two-stage detectors. Unfortunately, no counterparts to ROIs exist in single-stage detectors, making local adaptation much more challenging. This has only been tackled by~\cite{hsu2020every} for the specific detector of~\cite{tian2019fcos}, which explicitly extracts objectness maps, and by~\cite{CHEN_2021_I3NET}, which introduces complementary modules specifically designed for the SSD architecture~\cite{liu2016ssd}.


\begin{figure}[t]
\centering

\includegraphics[width=\linewidth]{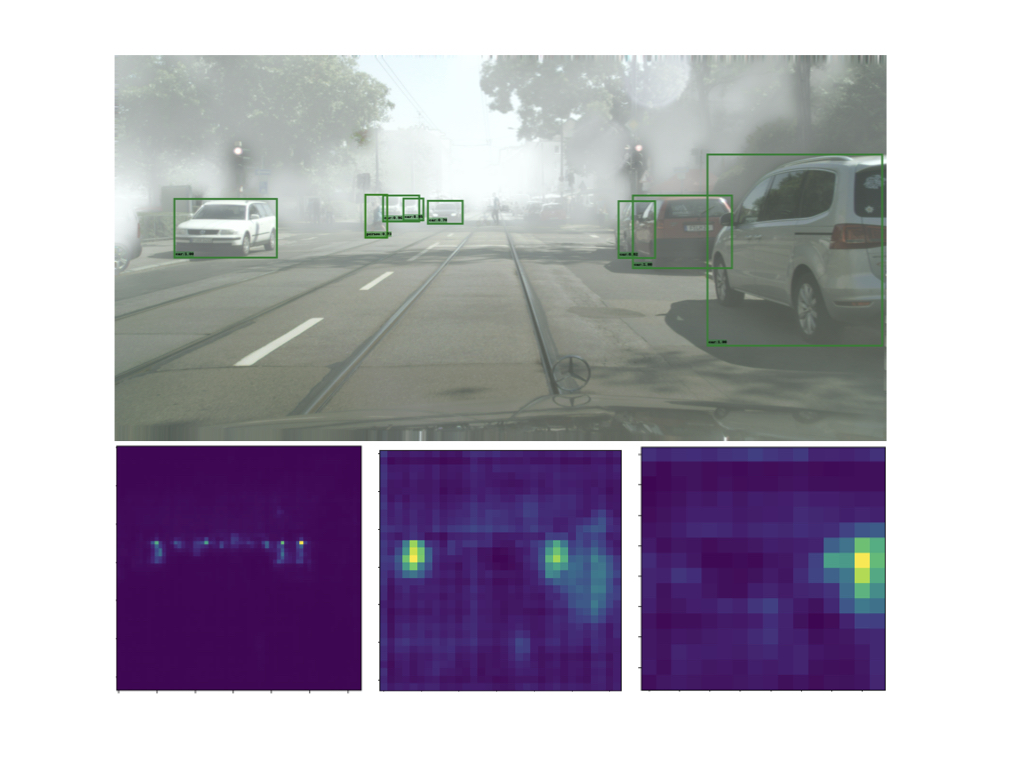}

\caption{{\bf Leveraging attention for local domain adaptation.} \textbf{Top:} Target image with predicted detections. \textbf{Bottom:} Attention maps output by our approach for feature maps at different scales, allowing us to focus adaptation on the relevant local image regions, ranging from small (left) to large (right) objects. The attention maps are re-scaled to the same size for visualization purpose. Best viewed digitally.
}
\label{d}
\end{figure}

In this paper, we introduce a domain adaptation strategy able to perform local adaptation while generalizing across single-stage object detectors. Specifically, we introduce an attention mechanism that allows adaptation to focus on the regions that matter for detection, that is, the foreground regions, as depicted by Fig.~\ref{d}. In essence, our approach leverages attention to perform local-level feature alignment, thus following the intuition that has proven successful in adapting two-stage detectors. Our attention mechanism is generic and can be incorporated into any single-stage detector. Furthermore, and contrarily to~\cite{hsu2020every} and~\cite{CHEN_2021_I3NET}, we gradually modulate the adaptation from global features to local features, which lets us give increasingly more importance to foreground features as training progresses. Consequently, this allows us to use the same domain classifiers for both global and local alignment, thereby leading to a simpler implementation than~\cite{hsu2020every} and~\cite{CHEN_2021_I3NET}.

We demonstrate the benefits of our approach via a series of experiments on several standard domain adaptation detection datasets. Despite its comparative simplicity, our method outperforms the state-of-the-art ones of~\cite{hsu2020every} and~\cite{CHEN_2021_I3NET}. Furthermore, our results evidence the generalizability of our domain adaptation strategy to different single-stage frameworks, including SSD~\cite{liu2016ssd} and YOLOv5~\cite{glenn_jocher_2020_4154370}, and the importance of local feature alignment over the global ones, particularly in the later training stages. We will make our code publicly available.


\section{Related Work}
\subsection{Object Detection}
Two-stage object detectors, such as FasterRCNN~\cite{ren2016faster}, consist of a feature extractor, a region proposal network (RPN) and a refinement network. The RPN provides foreground regions, via ROI pooling, to the refinement stage for bounding box prediction and classification. 
Recently, one-stage detectors~\cite{liu2016ssd,redmon2016you,glenn_jocher_2020_4154370,detr,tian2019fcos,lin2017focal,tan2020efficientdet} have emerged as an alternative, becoming competitive in accuracy while faster and more compact than two-stage ones. Most of them~\cite{liu2016ssd,glenn_jocher_2020_4154370,lin2017focal,tan2020efficientdet} rely on predefined bounding box anchors for prediction, and thus do not provide region proposals likely to contain foreground objects as two-stage detectors do. The only exception to this anchor-based approach to single-stage detection are the detectors of~\cite{tian2019fcos,detr}. Specifically,~\cite{tian2019fcos} yields an object centerness map, and~\cite{detr} learns object regions via a self-attention~\cite{vaswani2017attention} based encoder-decoder. Arguably, YOLO (\cite{redmon2016you},~\cite{glenn_jocher_2020_4154370}) predicts an objectness score for each anchor box, which could be leveraged to create an objectness map at the feature-level. However, we will show in section~\ref{sec:YOLOCOCO}, that our method is superior to this approach. In any event, in contrast to these approaches, we develop a self-attention framework for domain adaptation. It can be integrated into any anchor-based detector, which we illustrate using SSD~\cite{liu2016ssd} and YOLOv5~\cite{glenn_jocher_2020_4154370}.

\subsection{Domain Adaptation for Object  Detection}
While the bulk of the domain adaptation literature focuses on image classification, several works have nonetheless tackled the task of unsupervised domain adaptation for object detection. In particular, most of them have focused on the two-stage FasterRCNN detector. In this context,~\cite{chen2018domain} uses instance- and image-level alignment to improve the FasterRCNN performance on new domains; \cite{saito2019strong} shows that a strong local feature alignment improves adaptation, particularly when focusing on foreground regions; \cite{chen2020harmonizing} performs feature- and image-level adaptation on interpolated domain images generated using a CycleGAN~\cite{zhu2017unpaired}; \cite{zhu2019adapting} clusters the proposed object regions using $k$-means clustering and uses the centroids to do instance-level alignment; \cite{shen2019scl} introduces a method to improve the interaction between local and global alignment. In essence, all of these works leverage the RPN proposals to achieve a form of local feature alignment, showing the importance of focusing adaptation on the foreground features. Here, we follow a similar intuition but develop a method applicable to single-stage detectors, which do not rely on an RPN.

Only few works have tackled domain adaptation for single-stage detectors. In particular,~\cite{kim2019self} proposes a method to generate reliable pseudo labels to train the SSD detector on new domains. Pseudo labels, however, are orthogonal to our work; we focus on feature alignment, and while our approach could further benefit from pseudo labels, studying this goes beyond the scope of this paper. Therefore,~\cite{hsu2020every} and~\cite{CHEN_2021_I3NET} constitute the works closest to our approach. Specifically, \cite{hsu2020every} uses the object centerness maps predicted by the single-stage detector of~\cite{tian2019fcos} to perform local feature alignment. While effective, this approach is therefore restricted to this specific detector. Here, by contrast, we introduce a general approach to local feature alignment in single-stage detection. \cite{CHEN_2021_I3NET} designs a set of complementary modules, which help global- and local-level alignment in the dissimilar domain setting, implicitly learning foreground regions in the SSD architecture.They formulate their category alignment loss for target domain using the class probabilities of each anchor boxes. SSD, as used in~\cite{CHEN_2021_I3NET}, uses softmax-based normalized prediction for each anchor box whereas, YOLOv5 does multiclass prediction  using logistic classifiers. Hence, the approach in~\cite{CHEN_2021_I3NET} doesn't translate directly with the multiclass prediction framework of YOLOv5. By contrast, our approach is agnostic to the kind of detection head. Furthermore, we also learn foreground regions implicitly, but rely on a simpler, generalizable strategy, yet outperforms both the approaches of \cite{hsu2020every} and~\cite{CHEN_2021_I3NET}. Specifically, while~\cite{hsu2020every} and~\cite{CHEN_2021_I3NET} continuously aim to adapt the global and local features throughout the whole training process, we gradually modulate adaptation from the global to the local level. This lets us focus more strongly on the foreground regions and use the same domain classifiers for global and local adaptation.

\subsection{Self-Attention}
Our approach exploits self-attention (SA). SA was introduced in~\cite{vaswani2017attention} for natural language processing and has since then become increasingly popular in this field~\cite{devlin2018bert,brown2020language}. Recently, it has also gained popularity in computer vision, for both image recognition~\cite{ramachandran2019stand,bello2019attention,dosovitskiy2020image} and object detection~\cite{detr}. While other attention mechanisms have been proposed~\cite{hu2018squeeze,woo2018cbam,fu2019dual,wang2017residual}, they typically require more architectural changes than vanilla SA~\cite{vaswani2017attention}, which motivated us to rely on this strategy in our method. In any event, while self-attention has been employed for various tasks, our work constitutes the first apporach at directly exploiting the learned attentions for unsupervised domain adaptation of generic single-stage detectors.


\section{Method}
 Let us now introduce our attention-based domain adaptation strategy for single-stage detection. 
\begin{figure}[t]
	\includegraphics[width=\linewidth]{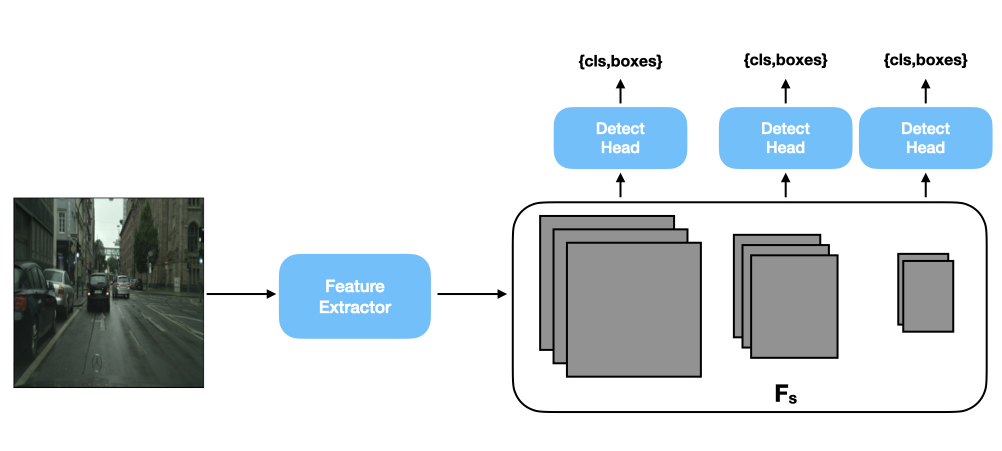}
	\caption{{\bf General single-stage object detection architecture.} Both SSD~\cite{liu2016ssd} and YOLOv5~\cite{glenn_jocher_2020_4154370}, used in our experiments, comply with this architecture, and other methods~\cite{lin2017focal,tan2020efficientdet} also do.}
	\label{fig:single-stage-arch}
\end{figure}

\subsection{Attention in Single-stage Detectors}
\label{sec:attention}
Single-stage object detectors typically follow the general architecture depicted by Figure~\ref{fig:single-stage-arch}, consisting of a feature extractor followed by several detection heads. These detection heads take as input the features ${F_s}$ at different scales $s\in[1,S]$, with the different scales allowing the detector to effectively handle objects of different sizes. Such an architecture directly predicts bounding boxes and their corresponding class from the feature maps, via the use of bounding box anchors at each spatial location. As such, it does not explicitly provide information about the features corresponding to the objects. This contrasts with two-stage detectors, whose region proposals directly correspond to potential objects.

\begin{figure}[t]
	\includegraphics[width=\linewidth]{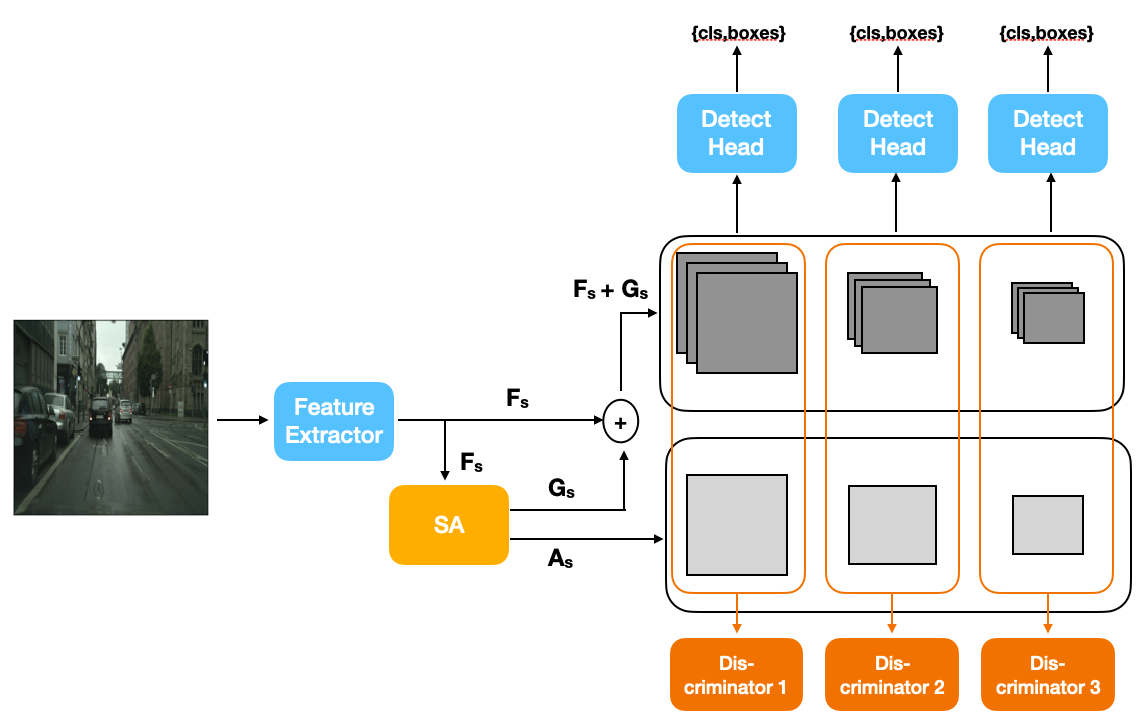}
	\caption{{\bf Overview of our approach.} We compute self-attention from the features extracted by the single-stage detector backbone. We then modulate these features with our attention maps so as to encourage the feature alignment achieved by the domain classifiers (discriminators) to focus on the relevant local image regions.}
	\label{fig:our-arch}
\end{figure}
To automatically extract information about the object locations, we propose to incorporate a self-attention mechanism~\cite{vaswani2017attention} in the detector. Intuitively, we expect the foreground objects to have higher self-attention than background regions because the detector aims to identify them, and thus exploit self-attention to extract an objectness map.  To this end, we use an attention architecture similar to that of~\cite{detr}, but without attention-based decoder because we want to keep the same detector heads as in~\cite{liu2016ssd} and~\cite{glenn_jocher_2020_4154370}.

The attention module takes as input the feature map $F_s \in \mathbb{R}^{H_s\times W_s \times C_s}$ and produces an objectness map $A_s \in \mathbb{R}^{H_s\times W_s}$ and a feature map $G_s \in \mathbb{R}^{H_s\times W_s \times C_s}$.
Specifically, $F_s$ is flattened to $\mathbb{R}^{H_sW_s\times C_s}$ and transformed into a query matrix $Q\in \mathbb{R}^{H_sW_s\times D}$, a key matrix $K\in \mathbb{R}^{H_sW_s\times D}$ and value matrix $V\in \mathbb{R}^{H_sW_s\times C_s}$ using three separate linear layers. We then compute, $ A^{\prime}_s  \in \mathbb{R}^{H_sW_s\times H_sW_s}\;$
 \begin{equation}
 A^{\prime}_s = softmax\left(\frac{QK^T}{\sqrt{D}}\right) 
 \end{equation}
which, intuitively, represents the similarity between the query and the key at different spatial locations. To compute the objectness map $A_s$, we first take the maximum of $A^{\prime}_s$ in the second dimension, reshape the resulting vector to $\mathbb{R}^{H_s\times W_s}$, and then min-max normalize the resulting matrix to the $[0,1]$ range.
Given $A^{\prime}_s$, we also compute $G^\prime_s \in \mathbb{R}^{H_sW_s\times C_s}$
 \begin{equation}
 G^\prime_s = A^{\prime}_sV
 \end{equation}
which we reshape to $\mathbb{R}^{H_s\times W_s \times C_s}$ to obtain $G_s$. We then pass $F_s+G_s$ to the detection head. 
In addition to this, and as will be discussed in more detail in Section~\ref{sec:uda}, we further leverage $A_s$ to modulate the $F_s+G_s$ features for domain adaptation. This differs from previous SA works, which do not explicitly exploit the learnt attention maps.


In practice, instead of the single-head attention mechanism discussed above, we rely on the multi-head extension presented in detail in~\cite{vaswani2017attention,detr}. In short, multi-head attention extracts different representations for the same pair of locations, and these representations are combined via feature concatenation followed by a linear projection.

As the different detection heads focus on objects of different sizes, we add an attention module at each scale. These modules are trained jointly with the feature extractor and detection heads. Because we do not have access to supervisory signal for the attention/objectness maps, the loss function $\mathcal{L}^{det}$ to train the detector remains the same as that of the original single-stage detector.
Typically~\cite{liu2016ssd,glenn_jocher_2020_4154370}, such a loss function incorporates a classification term to categorize pre-defined anchor bounding boxes, and a regression one to refine these anchors. It can thus be expressed in general as
\begin{align}
\mathcal{L}^{det}(I) = \mathcal{L}^{cls}(I) + \mathcal{L}^{reg}(I)\;.
\end{align}

\subsection{Unsupervised Domain Adaptation}
\label{sec:uda}
Let us now explain how we exploit the above-mentioned attention mechanism for unsupervised domain adaptation. This process is depicted by Fig.~\ref{fig:our-arch}. Let $I_{s}$ be a source image, for which we have the ground-truth bounding boxes and class labels, and $I_{t}$ be a target image, for which we do not. The source and target images are drawn from two different distributions but depict the same set of classes. Domain adaptation then translates to learning a representation that reduces the gap between both domains.

An effective approach to achieve this consists of jointly training a domain discriminator $D$ in an adversarial manner~\cite{ganin2016domain}, encouraging the learnt features not to carry any information about the observed domain. In our context, because the detection heads act on features at different scales, we use a separate discriminator $D_s$ for each scale $s$. However, we do not directly use the feature maps $F_s$ as input to these discriminators, but instead aim to focus the adaptation on the foreground objects, accounting for the fact that the background can genuinely differ across the two domains.

To this end, we leverage the objectness maps from Section~\ref{sec:attention} to extract the weighted feature map 
\begin{equation}
M_s = (1-\gamma)*(F_s+G_s) +\gamma*(F_s+G_s)\odot A_s\;,
	\label{eq:feats}
\end{equation}
where $\odot$ indicates an element-wise product performed independently for each channel of $(F_s+G_s)$, and $\gamma \in [0,1]$. 
This formulation combines the global, unaltered features with the local ones obtained by modulating the features by our attention map.
During our training, we then gradually increase $\gamma$ from 0 to 1, which lets us transition from global adaptation to local feature alignment. Intuitively, this accounts for the fact that, at the beginning of training, the predicted attention maps may be unreliable, and a global alignment is thus safer. We also observed such a strategy to facilitate the training of the discriminators. In practice, we compute $\gamma$ as
\begin{align}
\gamma = \frac{2}{1+\exp(-\delta\cdot r)}-1\;,
\label{eq:gammaterm}
\end{align}
where $\delta$ controls the smoothness of the change and $r = \frac{\textrm{current}\;\textrm{iteration}}{\textrm{max}\;\textrm{iteration}}$.

Given the attention-modulated features $M_s$ for each scale $s$, we then write the discriminator loss as
\begin{equation}
\begin{aligned}
\mathcal{L}^{dis}(I) = -\frac{1}{S}\sum_s t & \log(D_s(M_s))  \\ &+(1-t) \log(1-D_s(M_s)),
\label{eqn:dis_loss}
\end{aligned} 
\end{equation}
where $t=0$, resp. $t=1$, indicates that image $I$ is a source, resp. target image. 

During training, the discriminator aims to minimize $\mathcal{L}^{dis}$ while the feature extractor seeks to maximize it. 
To facilitate such an adversarial training process, we use the gradient reversal layer (GRL) of~\cite{ganin2016domain}. Hence, the overall loss function minimize{}d by the feature extractor for a source and a target image can be expressed as
\begin{align}
\mathcal{L}(I_{s}) &= \mathcal{L}^{det}(I_{s})-\mathcal{L}^{dis}(I_{s})\;, \\
\mathcal{L}(I_{t}) &= -\mathcal{L}^{dis}(I_{t})\;,
\end{align}
 respectively. Note that, unlike~\cite{hsu2020every} and~\cite{CHEN_2021_I3NET}, we do not use pixel-wise domain discriminators, as we found our attention-modulated feature maps to be sufficient to suppress the background features.	Moreover, the formulation in Eq.~\ref{eq:feats} allows us to use the same discriminator for global alignment in the beginning of training and local alignment in the later training stages.

\section{Experiments}
In this section, we discuss our experimental settings and analyze our results.



\begin{table*}
	\centering 
	\begin{tabular}{c c c c c c c c c c}
	\hline
	Method & person & car  & train  & rider  & truck  & motorcycle & bicycle & bus & mAP@0.5\\ 
	\hline 
	~\cite{hsu2020every} - w/o DA & 18 & 28.3  &1.6   & 18.3  & 6.5  & 6.6 & 15.5 & 16.5 &13.9   \\ 
	~\cite{hsu2020every} - global & 25.1  &43.3   &5.4   &27.6   & 17.8  & 11.9 &  22.1& 33.5 &23.3  \\ 
	~\cite{hsu2020every} -global+local & \textbf{26.6}  &44.5   &4.8   &26.2   &\textbf{21.2}   & 12.3 & 19.1  & 33.9  &23.5\\
	I$^3$Net~\cite{CHEN_2021_I3NET} & 19.7 & 37.9 & \textbf{9.6} & 22.9 & 12.5 & \textbf{18.3} & 22.7 & 21.1 & 20.6 \\
	SSD - w/o DA & 15.1 & 28.8  & 0.2  & 12.9   & 2.2  & 5.8& 13.7  &13.5  &11.5\\
		SSD + our DA & 23.4  & \textbf{49.1}  &4.9   &\textbf{27.8}   & 16.9  & 17.6 &\textbf{24.2} &\textbf{34.0}  &\textbf{24.8}\\ 
	\hline
	\end{tabular}
	\caption{Results on Cityscapes to Foggy adaptation}
	\label{tab:city2foggyssd}
\end{table*}

\subsection{Datasets}
We evaluate our method using the following four standard datasets:\\
\textbf{Cityscapes}~\cite{cordts2016cityscapes} contains 2975 images in the training set and 500 in the test set, with annotations provided for eight categories, namely, \emph{person, car, train, rider, truck, motorcycle, bicycle and bus}. The images depict street scenes taken from a car, mostly in good weather conditions. \\
\textbf{Foggy Cityscapes}~\cite{sakaridis2018semantic} contains synthetic images aiming to mimic the Cityscapes setting, but in foggy weather. It contains 2965 training images and 500  testing ones, depicting the same eight categories as Cityscapes. \\
\textbf{Sim10K}~\cite{johnson2016driving} consists of 9975 synthetic images, with annotations available for the \emph{car} category.\\
\textbf{KITTI}~\cite{geiger2012we} depicts street scenes similar to those of Cityscapes, but acquired using a different camera setup. In our experiments, we will only use its 6684 training images.

Following~\cite{hsu2020every}, we present results for the following domain adaptation tasks: \\
\textbf{Sim$\rightarrow$Cityscapes (\textbf{S$\rightarrow$C})}: This evaluates the effectiveness of a method to adapt from synthetic data to real images. All Sim10K images are used as source domain, and the Cityscapes training images act as target domain.  Following~\cite{hsu2020every}, only the \emph{car} class is considered for evaluation.\\
\textbf{KITTI$\rightarrow$Cityscapes (\textbf{K$\rightarrow$C})}: This task aims to evaluate adaptation to a different camera setup. We use the KITTI training images as source domain and the Cityscapes training images as target one. Again, as in~\cite{hsu2020every}, we consider only the \emph{car} class for evaluation.\\
\textbf{Cityscapes$\rightarrow$Foggy Cityscapes (\textbf{C$\rightarrow$F})}: The goal of this experiment is to test the effectiveness of a method in different weather conditions. We use the Cityscapes training images as source domain and all Foggy Cityscapes images as target data. For this task, all eight object categories are taken into account for evaluation.


\subsection{Implementation Details}

We evaluate our method on two single-stage detectors, SSD~\cite{liu2016ssd}\footnote{\url{https://github.com/lufficc/SSD}} and YOLOv5~\cite{glenn_jocher_2020_4154370}\footnote{\url{https://github.com/ultralytics/yolov5}}. We implemented our method in Pytorch, and performed all our experiments on a single V100 GPU. The batch consists of 8 images, 4 drawn from source and 4 from target domain. We set $\delta$ in Eq.~\ref{eq:gammaterm} to $5$. We provide additional training details in the supplementary material.

\textbf{SSD} relies on a similar VGG~\cite{simonyan2014very} backbone to that used by the detectors employed in~\cite{hsu2020every} and~\cite{CHEN_2021_I3NET}. We will therefore focus our comparison with~\cite{hsu2020every} and with~\cite{CHEN_2021_I3NET} to our SSD-based approach.  We employ an image resolution of $512\times512$ because it is the highest resolution available for the SSD architecture. Note that, in~\cite{hsu2020every}, larger images were used, i.e., a short image side between 800 and 1333, and that~\cite{CHEN_2021_I3NET} used a lower, $300\times300$ resolution. For the comparison to be fair, we thus re-trained these methods with this $512\times512$ image resolution.
To further make our SSD architecture comparable to that of~\cite{hsu2020every}, we incorporated a Feature Pyramid Network~\cite{lin2017feature} to our SSD backbone. 
Following~\cite{hsu2020every} and~\cite{CHEN_2021_I3NET}, all backbones were initialized with ImageNet-trained weights.

\begin{figure*}[t]
\centering
\begin{tabular}{cc}
\includegraphics[width=.45\linewidth]{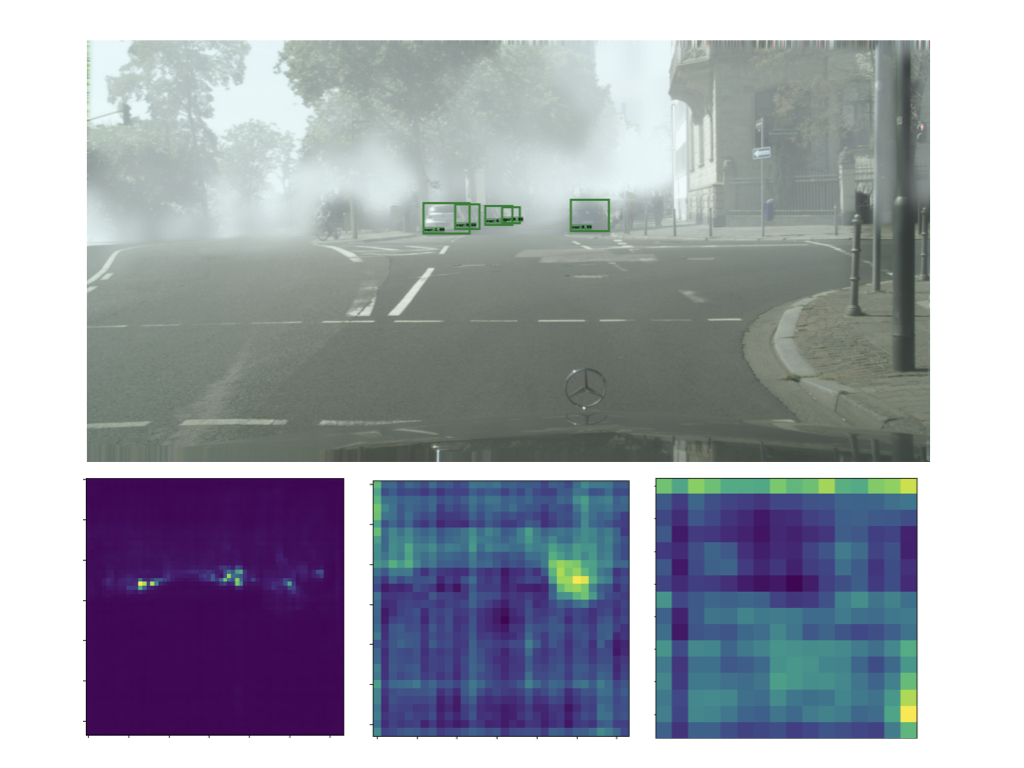}&
\includegraphics[width=.45\linewidth]{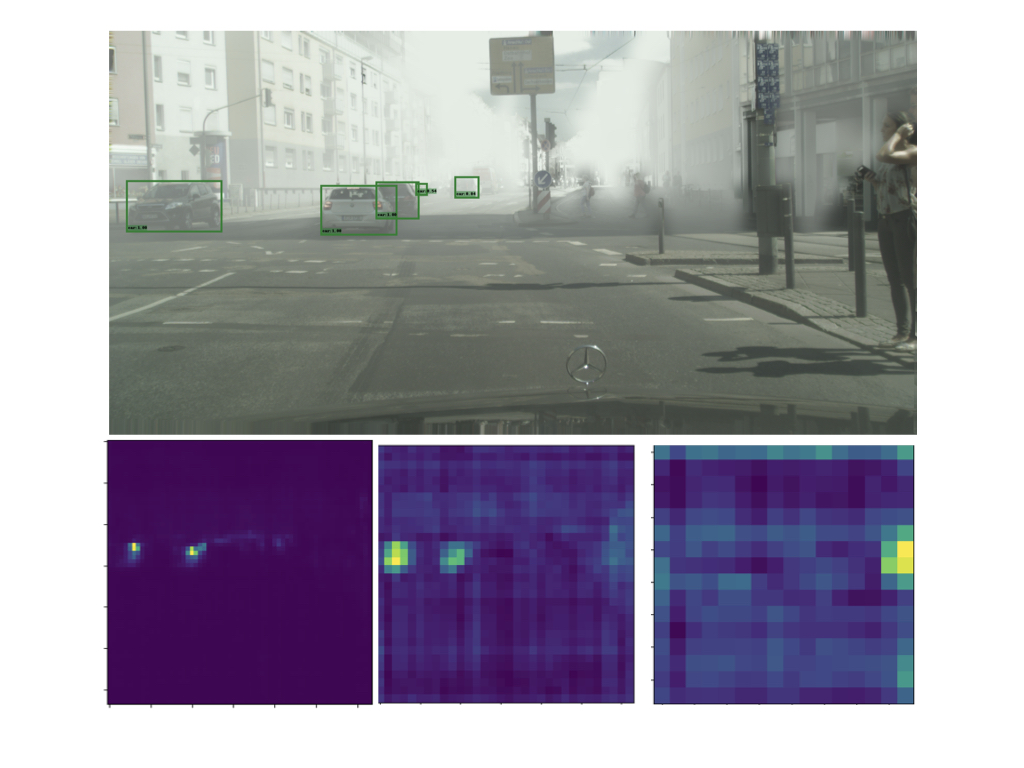}\\

\end{tabular}
  \caption{{\bf Qualitative results on \textbf{C$\rightarrow$F}.} We show target images with predicted detections, together with attention maps at different scales. While this adaptation task is particularly challenging, our attention maps nonetheless manage to correctly identify the objects at their different scales. Note, when there are no object of interest activation map tends to have activation everywhere. All predictions are with confidence 50\% and above.
 }
  \label{fig:attenmapsfoggy}
  \end{figure*}


\begin{figure*}[t]
\centering
\begin{tabular}{cc}
\includegraphics[width=.45\linewidth]{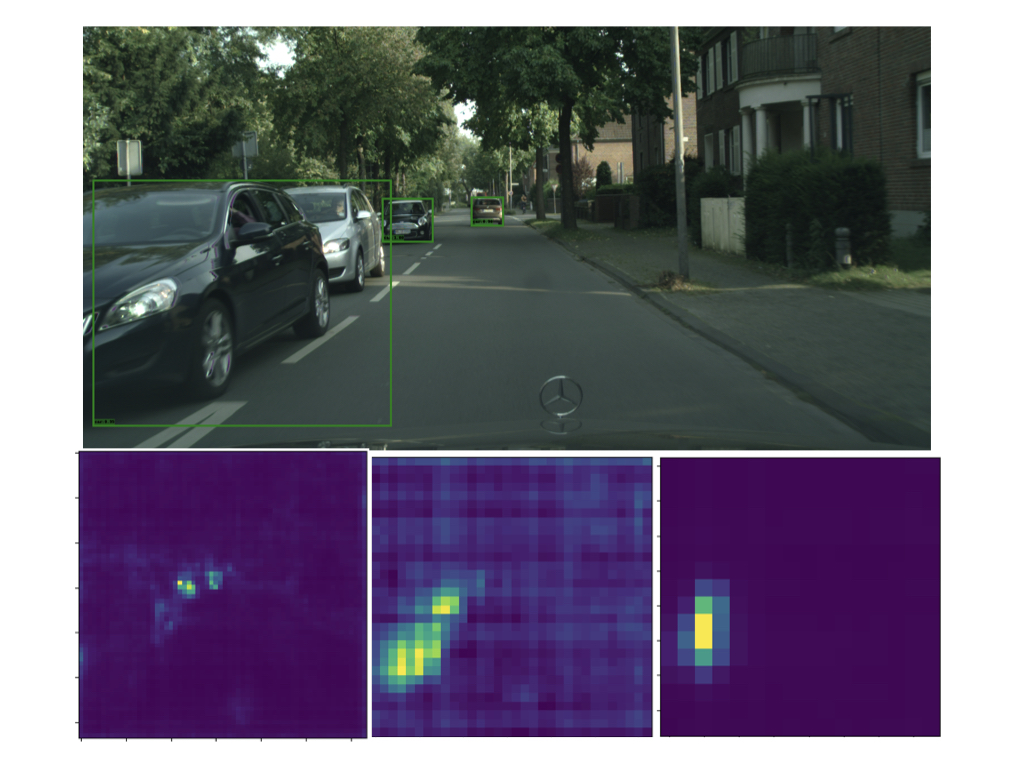}&
\includegraphics[width=.45\linewidth]{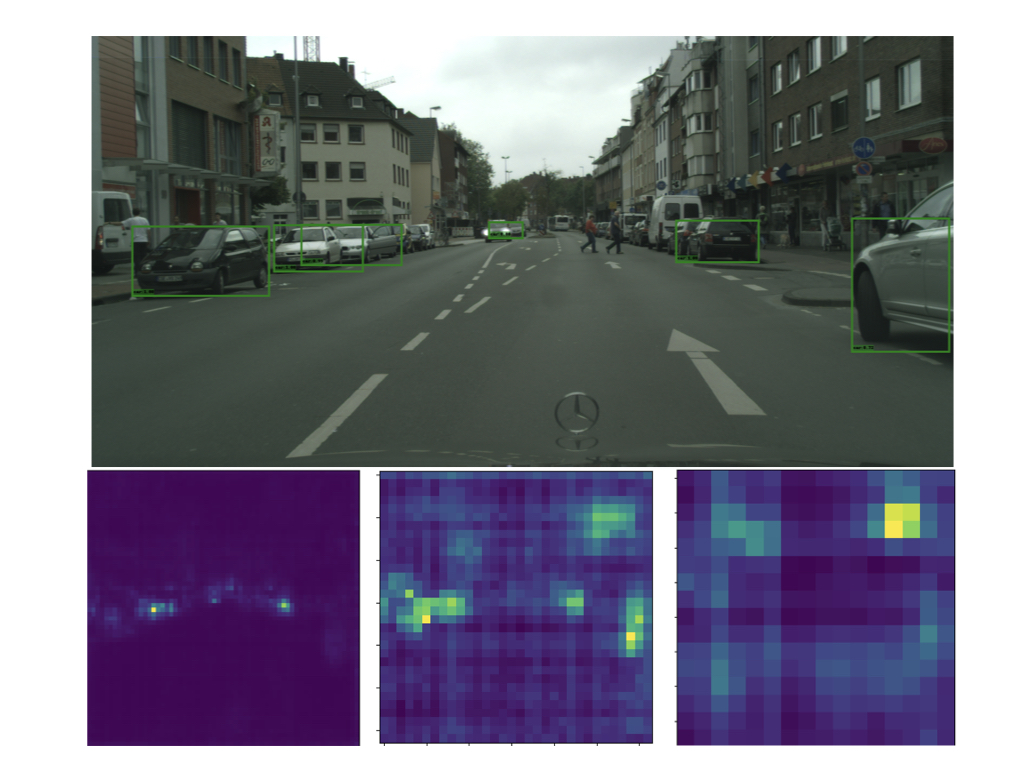}\\
\includegraphics[width=.45\linewidth]{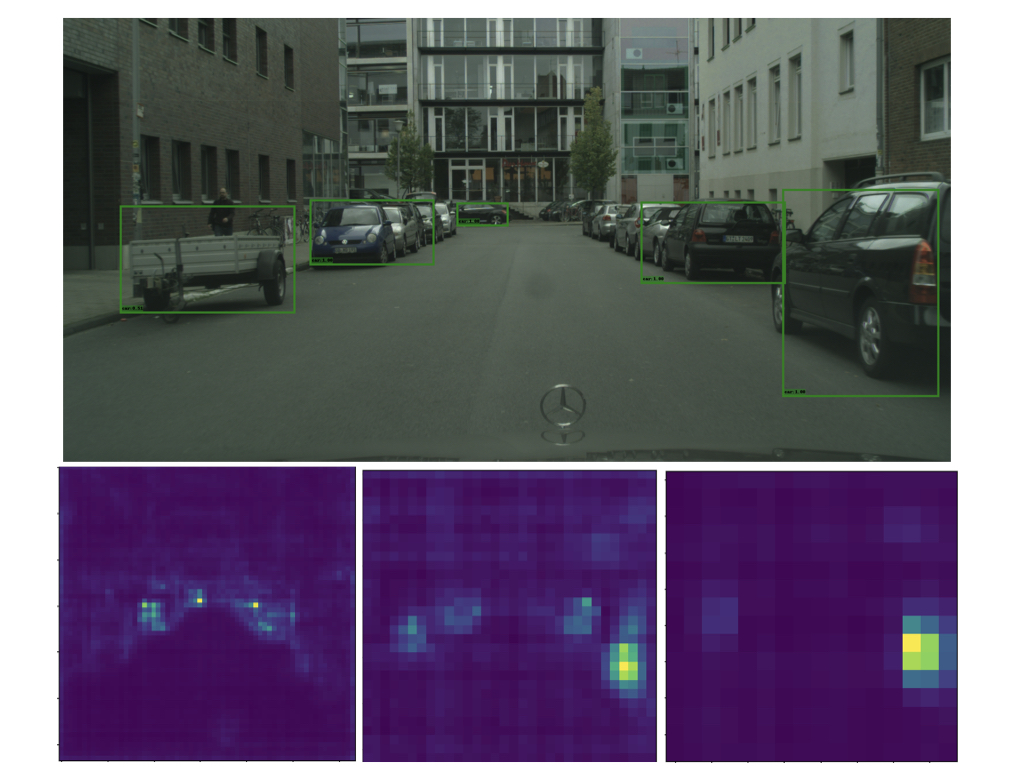}&
\includegraphics[width=.45\linewidth]{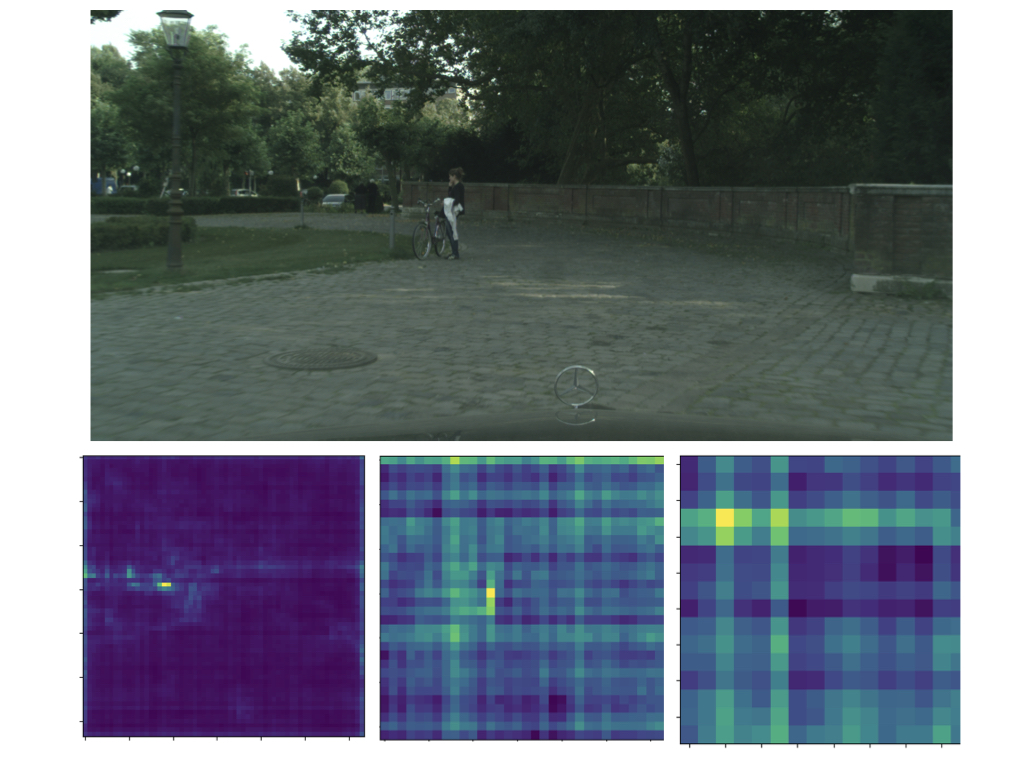}\\

\end{tabular}
  \caption{{\bf Qualitative results on \textbf{S$\rightarrow$C}.} We show target images with their predicted detections, together with the corresponding attention maps at different scales. Note that the finer map (left) correctly identifies the small cars whereas the coarser one (right) focuses on large cars. \textbf{Bottom right:} Because, this task focuses on cars only, this image does not contain any object of interest. Hence, in this case, the attention maps tend to have either no activation or activations everywhere. Note that the fine attention map nonetheless highlights cars in the background, which, by zooming in, can be verified to truly be present in the image. All predictions are with confidence 50\% and above.
 }
  \label{fig:attenmaps2c}
  \end{figure*}


\textbf{YOLOv5} is also trained with input images of size of $512\times 512$. This allows us to illustrate the generality of our approach to other single-stage detectors. Specifically, we use the YOLOv5s backbone, which is the smallest model out of all YOLO configurations. We keep the default configuration for preprocessing and data augmentation. We initialize the backbone with COCO-pretrained weights~\cite{lin2015microsoft} since~\cite{glenn_jocher_2020_4154370} don't provide ImageNet-trained weights. 

\subsection{Results}

\subsubsection{Comparison with the State of the Art}

Let us first compare our SSD-based method with \cite{CHEN_2021_I3NET} and with the global and local version of~\cite{hsu2020every}. \comment{Furthermore, we report the results of an additional baseline consisting of excluding the attention module from our backbone and performing domain adaptation directly on $F_s$. In other words, this baseline performs domain adaptation via a global feature alignment. We will refer to it as {\it SSD - w/o attn w DA}.} As a reference point, we also report the results without domain adaptation as {\it SSD - w/o DA}.



\begin{table}
	\centering 
	\begin{tabular}{c c}
	\hline
	Method & mAP@0.5\\
	\hline 
	~\cite{hsu2020every} - w/o DA & 31.5\\ 
	~\cite{hsu2020every} - global & 33\\ 
	~\cite{hsu2020every} - global+local & 32.8\\
	I$^{3}$Net~\cite{CHEN_2021_I3NET} & 35.1\\
	SSD - w/o DA & 29.1\\
	SSD + our DA & \textbf{36.7}\\ 
	\hline
	\end{tabular}
	\caption{Results on Sim10K to Cityscapes adaptation}
	\label{tab:sim2cityssd}
\end{table}


\begin{table}
	\centering 
	\begin{tabular}{c c}
	\hline
	Method & mAP@0.5\\
	\hline 
	~\cite{hsu2020every} - w/o DA & 33.3\\ 
	~\cite{hsu2020every} - global & 23.3\\ 
	~\cite{hsu2020every} - global+local & 27.8\\
	I$^3$Net~\cite{CHEN_2021_I3NET} & 40.0 \\
	SSD - w/o DA & 33.1\\
	SSD + our DA & \textbf{40.5}\\ 
	\hline
	\end{tabular}
	\caption{Results on KITTI to Cityscapes adaptation}
	\label{tab:kitti2cityssd}
\end{table}


\begin{table*}[t]
	\centering 
	\begin{tabular}{c c c c c c c c c c}
	\hline
	Method & person & car  & train  & rider  & truck  & motorcycle & bicycle & bus & mAP@0.5\\
	\hline 
	YOLO - w/o DA & 27.1  &40.8   & 4.5  & 30.8  & 11.1  & 9.3 & 21 & 24.7& 21.1\\
	YOLO + obj w DA &31.8 &50.3   & 4.9 &33.9   & 18.5  & \textbf{12.7} &\textbf{25.8} &34.3  & 26.5\\
		YOLO + our DA & \textbf{32.8} &\textbf{51.3}   &\textbf{16.2}   &\textbf{35.7}  &\textbf{18.8}   & 11.8 &25.6  & \textbf{34.5}  & \textbf{28.3}\\
	\hline
	\end{tabular}
	\caption{Results on Cityscapes to Foggy adaptation}
	\label{tab:city2foggyyolococo}
\end{table*}
Table~\ref{tab:city2foggyssd} provides the results on \textbf{C$\rightarrow$F}. Our method yields the best results on average (last column). When looking at the individual categories, we observe that we outperform all methods on {\it car}, {\it rider}, and additionally yield better results than~\cite{hsu2020every} on {\it bicycle}, with on par performance on {\it train} and {\it bus}. In some categories, such as {\it car}, our approach yields an increase in mAP by 10\% compared to~\cite{hsu2020every}. We attribute our poor performance on {\it train} and {\it truck} to the fact that these categories are under-represented in the source domain, and that their similar elongated shapes creates confusion between these classes. We outperform~\cite{CHEN_2021_I3NET} on most of the categories and increase the mAP score by 29.5\% and 61\% for {\it car} and {\it bus}, respectively. This shows the effectiveness of our method.

In Figure~\ref{fig:attenmapsfoggy}, we provide examples of detections and attention maps predicted with our approach on the \textbf{C$\rightarrow$F} task. Despite the challenging nature of this adaptation problem, \comment{as evidenced by Figure~\ref{fig:sample_domain_city_foggy},} our method correctly highlights the objects in the scene. The attention maps at different scales focus on objects of different sizes. We show additional qualitative results on pre and post adaptation in the supplementary material.

Table~\ref{tab:sim2cityssd} shows the results for the \textbf{S$\rightarrow$C} adaptation. Our method again yields the best results, outperforming both ~\cite{hsu2020every} and~\cite{CHEN_2021_I3NET}. Surprisingly, the global alignment of~\cite{hsu2020every} yields better performance than when further exploiting their local alignment. This suggests that both should not be given equal importance as training progresses. \comment{Our method also outperforms our baseline without any attention, hence validating the importance of accounting for the foreground regions during feature alignment.}

In Figure~\ref{fig:attenmaps2c}, we provide qualitative results for the \textbf{S$\rightarrow$C} task. These results evidence that the attention maps we produce correctly focus on the local regions of interest, i.e., the cars in this case. Furthermore, the maps at different scales account for objects at different sizes. We show additional qualitative results on pre and post adaptation in the supplementary material.

We provide the \textbf{K$\rightarrow$C} results in Table~\ref{tab:kitti2cityssd}. Note that the method of~\cite{hsu2020every} fails to adapt to the target data, yielding worse performance than their own no-DA baseline. This difference compared to the results provided in~\cite{hsu2020every} arises from the use of a smaller image size here, as discussed above. Note, however, that the fact that the {\it \cite{hsu2020every} - w/o DA} baseline, which we also re-trained, yields essentially the same performance as our {\it SSD - w/o DA} baseline, and that the method of~\cite{hsu2020every} yields reasonable performance in the other source-target pairs evidence that we correctly re-trained this model. For this adaptation task, we achieve comparable results with~\cite{CHEN_2021_I3NET} even though we adopt simpler training and architecture choices. We show the qualitative results on this task in the supplementary material.

\subsubsection{Generalization to Another Architecture}
\label{sec:YOLOCOCO}

To show the generality of our approach, we use it with the YOLOv5 detector. We compare our method with an additional baseline \emph{YOLO + obj w DA}. This baseline leverages the fact that the YOLO architecture predicts an objectness score for each anchor box at each feature map location. We thus use the maximum score at each location to create an objectness map and replace our  $A_s$, learned using self-attention, with this map. Furthermore, we provide the results of the YOLOv5 architecture without domain adaptation as \emph{YOLO w/o DA}. 


The results on \textbf{C$\rightarrow$F}, \textbf{S$\rightarrow$C}, and \textbf{K$\rightarrow$C}  are shown in Table~\ref{tab:city2foggyyolococo}, Table~\ref{tab:sim2cityyolococo}, and Table~\ref{tab:kitti2city_yolococo}, respectively. As in the SSD case, our method consistently outperform the baselines, illustrating the generality of our approach. \emph{YOLO + obj w DA} performs worse than us on \textbf{S$\rightarrow$C}, \textbf{C$\rightarrow$F} and comparably on \textbf{K$\rightarrow$C}. This further shows that our attention scheme helps to learn better objectness maps.



\begin{table}
	\centering 
	\begin{tabular}{c c }
	\hline
	Method & mAP@0.5\\
	\hline
	YOLO - w/o DA & 42.5 \\
	YOLO + obj w DA & 43.5\\
	YOLO + our DA & \textbf{44.9}\\
	\hline
	\end{tabular}
	\caption{Results on Sim10K to Cityscapes adaptation}
	\label{tab:sim2cityyolococo}
\end{table}

\begin{table}
	\centering 
	\begin{tabular}{c c }
	\hline
	Method & mAP@0.5\\
	\hline
	YOLO - w/o DA & 29.1 \\
	YOLO + obj w DA & 37.5\\
	YOLO + our DA & \textbf{37.7}\\
	\hline
	\end{tabular}
	\caption{ Results on KITTI to Cityscapes adaptation}
	\label{tab:kitti2city_yolococo}
\end{table}

\subsection{Ablation Study}

\subsubsection{Global vs Local Alignment}

As mentioned in Section~\ref{sec:uda}, our formulation in Eq.~\ref{eq:feats} is motivated by the intuition that one should initially perform a global alignment to learn reliable features for the attention module, but that the global features can be gradually dropped to focus on local regions in the later training stages. To further evaluate the benefits of local vs global alignment, we implemented three alternative strategies: \textbf{(a)} The global features are maintained throughout the whole training process. Concretely, this strategy computes a features map of the form
\begin{align}
M_d = (F_s+G_s) +\gamma*(F_s+G_s)\odot A_s\;,
\label{eq:m_d}
\end{align} 
where $\gamma$ follows the same rule as in our approach. 
\textbf{(b)} We set $\gamma = 1$ in Eq.~\ref{eq:feats}, which corresponds to performing adaptation using only local features throughout the whole training process. \textbf{(c)} We set $\gamma = 0$ in Eq.~\ref{eq:feats}, which corresponds to a global alignment where the attention block is nonetheless employed via $G_s$ but the attention maps are not used to modulate the features.


\begin{table}[h]
	\centering 
	\begin{tabular}{ c c }
	\hline
	Method & mAP@0.5\\
	\hline
	 Ours w. Eq.~\ref{eq:m_d} & 32.9 \\
	Ours w. $\gamma$ =1 & 33.6 \\
	Ours w. $\gamma$ = 0 & 34.2 \\
	Ours &\textbf{36.7}\\
	\hline
	\end{tabular}
	\caption{Global vs Local Alignment on \textbf{S}$\rightarrow$\textbf{C}}
	\label{tab:global_local}
\end{table}

As shown in Table~\ref{tab:global_local} for the \textbf{S}$\rightarrow$\textbf{C} task and with an SSD-based detector, our approach outperforms all of these baselines. This confirms that maintaining a global alignment term throughout training harms the overall performance, suggesting that the transition from global to local is crucial. This is further supported by the fact that local or global alignment on their own perform better than combining both in a suboptimal fashion. Purely local adaptation yields worse results than purely global adaptation because the attention maps do not carry sufficient meaningful information at the beginning of training, which compromises the rest of the training process. This study shows that both global and local alignment are important, and that their interaction affects the overall performance.

\subsubsection{Hyperparameter Study}
\begin{figure}[t]
\centering
\begin{tabular}{c}
\includegraphics[width=2.5in]{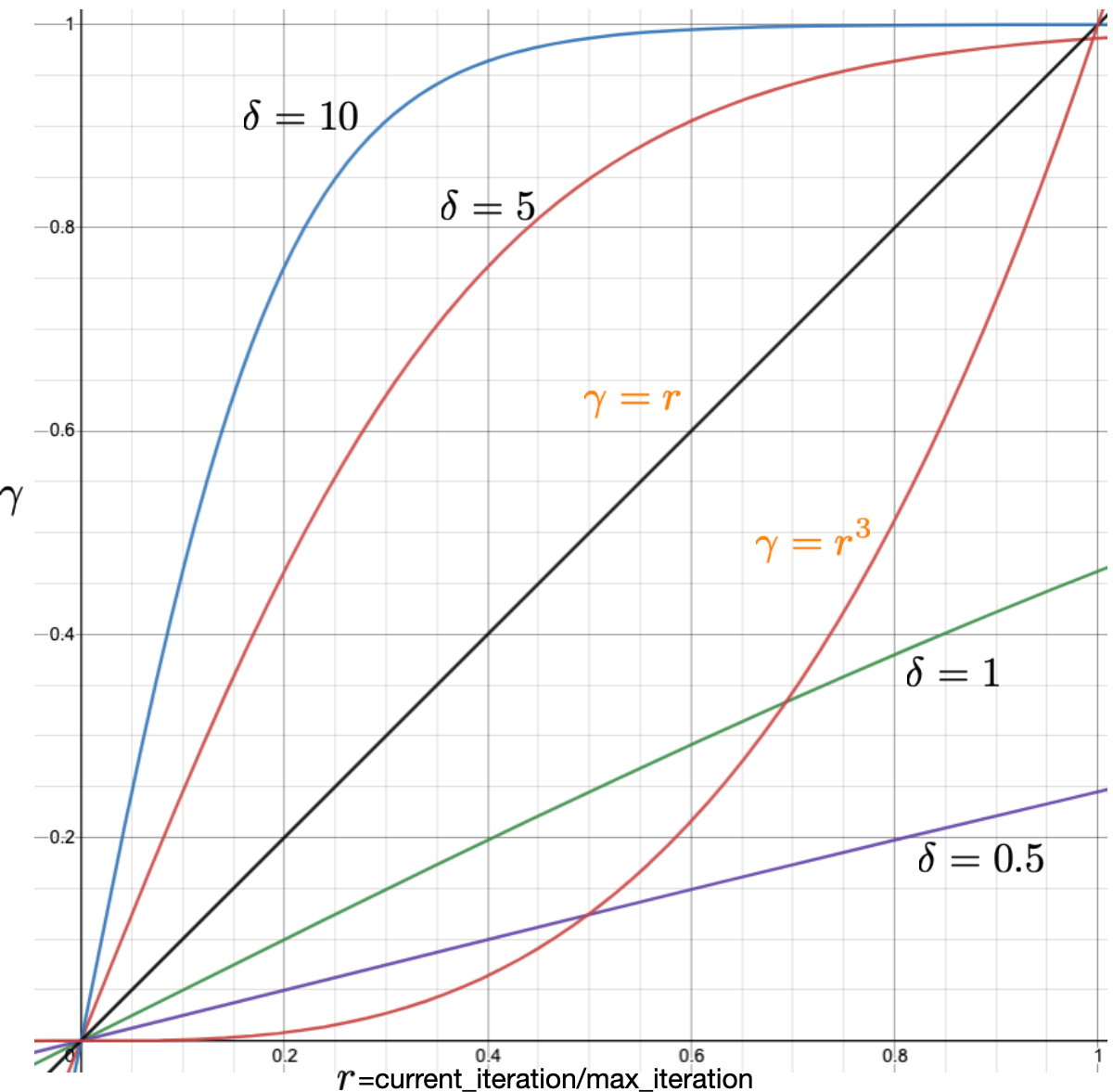}
\end{tabular}
  \caption{{\bf Study of different variants of $\gamma$.} We plot the evolution of $\gamma$ throughout training for different values of $\delta$.  We also study other functions highlighted in \textcolor{orange}{orange}.} 

  \label{fig:gammavshyp}
  \end{figure}

In this section, we further investigate the influence of attention on our results. To this end,
we first study the effect of $\delta$ in  $\gamma = \frac{2}{1+\exp(-\delta\cdot r)}-1$ for {\bf S$\rightarrow$ C} with SSD. Table~\ref{tab:delta_gamma} shows mAP scores for strategies ranging from local alignment ($\gamma$=1) to more global alignment ($\delta$=0.5). Figure~\ref{fig:gammavshyp} depicts the evolution of $\gamma$ for different values of $\delta$. For $\delta= 10, 5$ we see that the transition from global to local is relatively fast, which yields better results than the slower transition $\delta=1,0.5$ and  $\gamma$=$r^3$. We attribute this to the fact that the network becomes biased towards global features if the transition is slow. Moreover, for $\delta=1,0.5$, the local features are never given much importance as $\gamma$ is always below 0.5. Finally, we see that a linear function $\gamma $= $r$ yields a similar score to that obtained with a non-linear function with $\delta=10$, suggesting that transition leads to a better result, thereby validating our claim of the importance of global adaptation in the initial training stages and local adaptation towards the end.


\begin{table}[h]
	\centering 
	\begin{tabular}{c c}
	\hline
	Method & mAP@0.5 \\
	\hline	 
	Ours w. $\gamma$ =1; large $\delta$ & 33.6 \\
	Ours w. $\delta$=10 & 35.6 \\
	Ours w. $\delta$=5 & \textbf{36.7}\\
	Ours w. $\gamma$ = $x$ & 35.7 \\
	Ours w. $\gamma$ = $x^3$ & 33.0 \\
	Ours w. $\delta$=1 & 33.4 \\
	Ours w. $\delta$=0.5 & 33.3 \\
	\hline
	\end{tabular}
	\caption{Hyperparameter study on \textbf{S}$\rightarrow$\textbf{C}}
	\label{tab:delta_gamma}
\end{table}



   

\begin{table}
	\centering
  \begin{tabular}{c c c }
    \hline
    Method  & \texttt{SSD} & \texttt{YOLO}  \\
    \hline
    w/o attn &  35.1 & 42.7  \\
    attn & \textbf{36.7} & \textbf{44.9}  \\
    \hline
  
  \end{tabular}
  \caption{{\bf Importance of Attention:} We show on \textbf{S}$\rightarrow$\textbf{C} the effectiveness of our attention mechanism. We report the mAP@0.5 on the target domain.}
	\label{tab:woattn}
\end{table}

\subsubsection{Importance of Attention}
To show the importance of attention, we trained both the SSD and YOLO detectors on \textbf{S}$\rightarrow$\textbf{C} without and with attention mechanism, along with domain adversarial training. As shown in Table~\ref{tab:woattn}, including our attention scheme does help to improve the performance on the target domain. 


\section{Conclusion}
To conclude, we have proposed to incorporate an attention module acting on the features extracted by the detector backbone, and to modulate these features so as to focus adaptation on the local foreground image regions that truly matter for detection. We have further developed a gradual training strategy that smoothly transitions from global to local feature alignment. Our experiments on several domain adaptation benchmarks have demonstrated that (i) with a comparable architecture, our method outperforms the state-of-the-art domain adaptation techniques for single-stage detection, despite the fact that they were designed for specific architectures; (ii) our approach remains effective across different single-stage detectors; (iii) our gradual training strategy effectively allows the network to benefit from global and local adaptation. In the future, we will study the use of pseudo labels with our local feature alignment strategy.

{\small
\bibliographystyle{ieee_fullname}
\bibliography{egbib}
}

\end{document}